\documentclass[letterpaper, 10 pt, conference]{ieeeconf}
\IEEEoverridecommandlockouts
\let\labelindent\relax
\usepackage{graphicx} \graphicspath{ {figure/} }
\usepackage{amsmath,amssymb,mathabx}
\usepackage[cal=cm]{mathalfa}
\usepackage{algorithmic}
\usepackage[ruled]{algorithm2e}
\usepackage{acronym}
\usepackage{enumitem}
\usepackage{hyperref}
\usepackage{balance}
\usepackage{xspace,setspace}
\usepackage[skip=3pt,font=small]{subcaption}
\usepackage[skip=3pt,font=small]{caption}
\usepackage[misc]{ifsym}
\usepackage[dvipsnames]{xcolor}
\usepackage[capitalise]{cleveref}
\usepackage{tabularx,multirow,array,makecell}
\usepackage{overpic}

\makeatletter
\DeclareRobustCommand\onedot{\futurelet\@let@token\@onedot}
\def\@onedot{\ifx\@let@token.\else.\null\fi\xspace}
\def\eg{\emph{e.g}\onedot} 
\def\ie{\emph{i.e}\onedot}

\def\sarw{shared AR workspace}
\makeatother

\makeatletter
\def\BState{\State\hskip-\ALG@thistlm}
\makeatother

\makeatletter
\renewcommand{\paragraph}{%
  \@startsection{paragraph}{4}%
  {\z@}{0ex \@plus 0ex \@minus 0ex}{-1em}%
  {\hskip\parindent\normalfont\normalsize\bfseries}%
}
\makeatother

\crefname{algocf}{alg.}{algs.}
\Crefname{algocf}{Algorithm}{Algorithms}

\acrodef{ar}[AR]{Augmented Reality}
\acrodef{vr}[VR]{Virtual Reality}
\acrodef{hri}[HRI]{Human-Robot Interaction}
\acrodef{fov}[FoV]{Field-of-View}
\acrodef{vpt}[VPT1]{Level 1 Visual Perspective Taking}

\title{\LARGE \bf Human-Robot Interaction in a Shared Augmented Reality Workspace}

\author{Shuwen Qiu$^{*}$\quad{}Hangxin Liu$^{*}$\quad{}Zeyu Zhang\quad{}Yixin Zhu\quad{}Song-Chun Zhu\quad{}
\thanks{$^{\star}$ Shuwen Qiu and Hangxin Liu contributed equally to this work.}
\thanks{UCLA Center for Vision, Cognition, Learning, and Autonomy (VCLA) at Statistics Department. Emails:
        {\tt\fontsize{8}{8}\selectfont\{s.qiu, hx.liu, zeyuzhang, yixin.zhu\}@ucla.edu}, \tt\fontsize{8}{8}\selectfont{sczhu@stat.ucla.edu}.}%
\thanks{The work reported herein was supported by ONR N00014-19-1-2153, ONR MURI N00014-16-1-2007, and DARPA XAI N66001-17-2-4029.}%
}

\begin{document}

\maketitle
\thispagestyle{empty}
\pagestyle{empty}

\begin{abstract}
We design and develop a new \emph{shared \ac{ar} workspace} for \ac{hri}, which establishes a bi-directional communication between human agents and robots. In a prototype system, the \sarw{} enables a \emph{shared perception}, so that a physical robot not only perceives the virtual elements in its own view but also infers the utility of the human agent---the cost needed to perceive and interact in \ac{ar}---by sensing the human agent's gaze and pose. Such a new \ac{hri} design also affords a \emph{shared manipulation}, wherein the physical robot can control and alter virtual objects in \ac{ar} as an active agent; crucially, a robot can proactively interact with human agents, instead of purely passively executing received commands. In experiments, we design a resource collection game that qualitatively demonstrates how a robot perceives, processes, and manipulates in \ac{ar} and quantitatively evaluates the efficacy of \ac{hri} using the \sarw{}. We further discuss how the system can potentially benefit future \ac{hri} studies that are otherwise challenging.
\end{abstract}

\section{Introduction}

Recent advance in \ac{vr} and \acf{ar} has blurred the boundaries between the virtual and the physical world, introducing a new dimension for \acf{hri}. With new dedicated hardware~\cite{lavalle2014head,liu2017glove,liu2019high}, \ac{vr} affords easy modifications of the environment and its physical laws for \ac{hri}; it has already facilitated various applications that are otherwise difficult to conduct in the physical world, such as psychology studies~\cite{schatzschneider2016turned,ye2017martian,wang2018spatially} and AI agent training~\cite{lin2016virtual,shah2018airsim,xie2019vrgym,xie2019vrgrasp}.

In comparison, \ac{ar} is not designed to alter the physical laws. By overlaying symbolic/semantic information and visual aids as holograms, its existing applications primarily focus on assistance in \ac{hri}, \eg, interfacing~\cite{weisz2017assistive,zhang2019vision,zhang2020congestion}, data visualization~\cite{collett2006augmented,ghiringhelli2014interactive,walker2018communicating}, robot control~\cite{kruckel2015intuitive,zolotas2018head}, and programming~\cite{liu2018interactive,quintero2018robot}. Such a confined range of applications hinders its functions in broader fields.

\begin{figure}[t!]
    \centering
    \includegraphics[width=\linewidth,trim={2cm 0.5cm 5cm 2.5cm},clip]{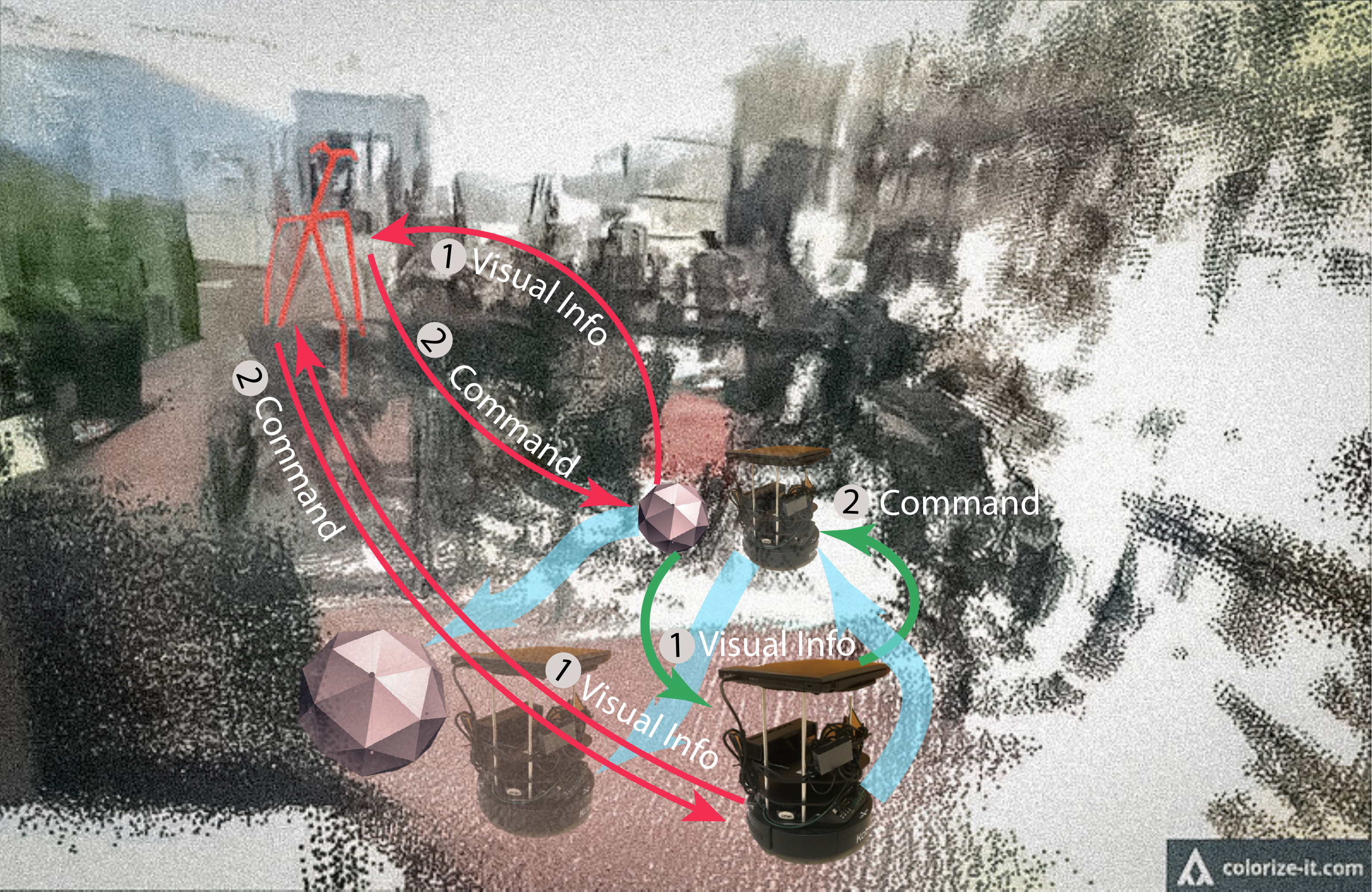}
    \caption{\textbf{A comparison between the existing \ac{ar} systems and the proposed \sarw{}.} Existing \ac{ar} systems limits to an \emph{active human, passive robot}, one-way communication, wherein a physical robot would only react to human commands via \ac{ar} devices without taking its own initiatives; see the red arrows. The proposed \sarw{} constructs an \emph{active human, active robot}, bi-directional communication channel that allows a robot to perceive and proactively manipulate holograms as human agents do; see green arrows. By offering shared perception and manipulation, the proposed \sarw{} affords more seamless \ac{hri}.}
    \label{fig:motivation}
\end{figure}

We argue such a deficiency is due to the current setting adopted in prior \ac{ar} work; we call it a \emph{active human, passive robot} paradigm, as illustrated by the red arrows in \cref{fig:motivation}. In such a paradigm, the virtual holograms displayed in \ac{ar} introduce asymmetric perceptions to humans and robots; from two different views, the robot and human agents may possess a different amount of information. This form of information asymmetry prevents the robot from properly assisting humans during collaborations. This paradigm also heavily relies on a one-way communication channel, which intrinsically comes with a significant limit: only human agents can initiate the communication channel, whereas a robot can only passively execute the commands sent by humans, incapable of proactively manipulating and interacting with the augmented and physical environment.

To overcome these issues, we introduce a new \emph{active human, active robot} paradigm and propose a \sarw{}, which affords shared perception and manipulation for both human agents and robots; see \cref{fig:motivation}:
\begin{enumerate}[leftmargin=*,noitemsep,nolistsep]
    \item \textbf{Shared perception} among human agents and robots. In contrast to existing work in \ac{ar} that only enhances human agents' understanding of robotic systems, the \sarw{} dispatches perceptual information of the augmented environment to both human agents and robots equivalently. By sharing the same augmented knowledge, a robot can properly assist its human partner during \ac{hri}; the robot can accomplish a \ac{vpt} by inferring if a human agent perceives certain holograms and estimating associated costs.
    \item \textbf{Shared manipulation} on \ac{ar} holograms. In addition to manipulating physical objects, \sarw{} endows a robot with the capability to manipulate holograms proactively, in the same way as a human agent does, which would instantly trigger the update of shared perception. As a result, \ac{hri} in the \sarw{} permits a more seamless and harmonious collaboration.
\end{enumerate}

\begin{figure*}[t!]
    \centering
    \includegraphics[width=\linewidth]{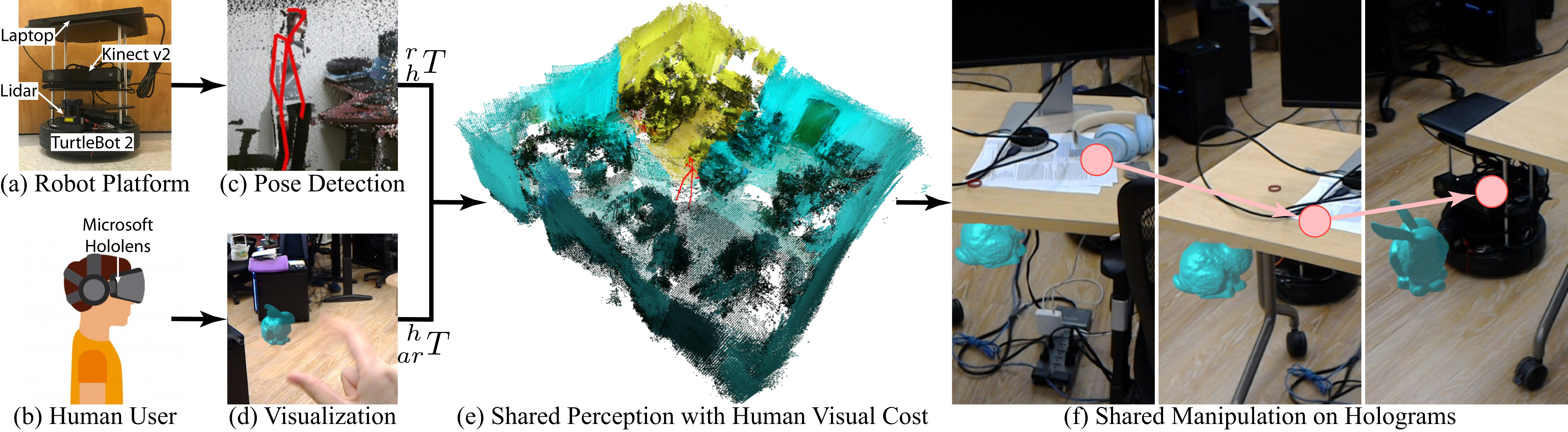}
    \caption{\textbf{A prototype system that demonstrates the concept of \sarw{}.} (a) A mobile robot platform with an RGB-D sensor and a Lidar for perception. (b) A human agent with an \ac{ar} headset (Microsoft HoloLens). By calculating (c) the transformation from the robot to the human, $^{r}_{h}T$, by a 3D human pose detector and (d) the transformation from the human to holograms, $^{~h}_{ar}T$, provided by the \ac{ar} headset, (e) the poses of holograms can be expressed in the robot's coordinate. Via \ac{vpt}, the robot estimates the utility/cost of a human agent to interact with a particular hologram: the yellow, light blue, and dark blue regions indicate where \ac{ar} holograms are directly seen by a human agent, seen after changing view angles, and occluded, respectively. (f) The system also endows the robot the ability to manipulate the augmented holograms and update the shared perception, enabling more seamless \ac{hri} in \ac{ar}.}
    \label{fig:system_diagram}
\end{figure*}

We develop a prototype system using a Microsoft Hololens and TurtleBot2, and demonstrates the efficacy of the \sarw{} in a case study of a resource collection game.

The remainder of the paper is organized as follows. \cref{sec:setup} introduces the system setup and details some critical system components. The two essential functions, shared perception and shared manipulation of the proposed \sarw{}, are described in \cref{sec:workspace}. \cref{sec:exp} demonstrates the efficacy of the proposed system by a case study, and \cref{sec:related_work} concludes the paper with discussions on some related fields the system could potentially promote.

\section{System Setup}\label{sec:setup}

In this section, we describe the prototype system that demonstrates the concept of the \sarw{}; \cref{fig:system_diagram} depicts the system architecture. Our prototype system assumes (i) a human agent wearing an \ac{ar} device and (ii) a robot with perception sensors; however, the system should be able to scale up to multi-human, multi-robot settings.

\paragraph*{Robot Platform}
We choose TurtleBot2 mobile robot with a ROS compatible laptop as the robot platform; see \cref{fig:system_diagram}a. The robot's perception module includes a Kinect 2 RGB-D sensor and a Hokuyo Lidar, which constructs the environment's 3D structure using RTAB-Map~\cite{labbe2014online}. Once the map is built, the robot only needs to localize itself within the map by fusing visual and wheel odometry.

\paragraph*{AR Headset}
Human agents in the present study wear a Microsoft HoloLens as the \ac{ar} device; see \cref{fig:system_diagram}b. HoloLens headset integrates a 32-bit Intel Atom processor and runs Windows 10 operating system onboard. Using Microsoft's Holographic Processing Unit, the users can realistically view the augmented contents as holograms. The \ac{ar} environment is created using the Unity3D game engine.

\paragraph*{Communication}
Real-time interactions in the \sarw{} demands timely communication between HoloLens (human agents) and robots, established using ROS\#~\cite{ros_sharp}. Between the two parties, HoloLens serves as the client, who publishes the poses of holograms, whereas the robot serves as the server, which receives these messages and integrates them into ROS. In addition to the perceptual information obtained by its sensors, the robot also has access to the 3D models of holograms so that they can be rendered appropriately and augmented to the shared perception.

\paragraph*{Overall Framework}
The shared perception in the \sarw{} allows a robot to perceive virtual holograms in three different levels with increasing depth: (i) know the existence of holograms in the environment, (ii) see the holograms from the robot's current coordinate obtained by localizing itself using physical sensors, and (iii) infer human agent's utility/cost of seeing holograms. Take an example shown in \cref{fig:system_diagram}e: human agents can directly see objects in the yellow region as it is within their \ac{fov}, but they need to change the views to perceive the objects marked in light blue; objects in dark blue are fully occluded. Only having with such a multi-resolution inference could the robot properly initiate interactions or collaboration with the human, forming a bi-directional communication. For instance, in \cref{fig:system_diagram}f, the robot estimates a hologram is occluded from the human agent's current view and plans and carries this occluded hologram to assist a human agent to accomplish a task. Since the robot proactively helps the human agent form collaborations, such a new \ac{ar} paradigm contrasts the prior one-directional communication.

\section{Shared AR Workspace}\label{sec:workspace}

Below we describe the shared perception and shared manipulation implemented in the \sarw{}.

\subsection{Detection and Transformation} 

A key feature in the \sarw{} is the ability to know where the holograms are at all time, which requires to localize human agents, robots, and holograms, and construct transformations among them. Using an \ac{ar} headset, the human agent's location is directly obtained. Given the corresponding transformations between a human agent and a hologram $i$, $_{i}^{h}T$, the \ac{ar} headset with the human agent's egocentric view can render the holograms.

By estimating the human pose from a single RGB-D image~\cite{zimmermann20183d}, the robot establishes a transformation to the human agent $_{h}^{r}T$; \cref{fig:system_diagram}c shows one example. Specifically, the frame of a human agent is attached to the head, whose $x$ axis is aligned with the human face's orientation estimated by three key points---two eyes and the neck. When the human agent is partially or completely outside of the robot's \ac{fov}, the frame of the human agent is directly estimated by leveraging the visual odometry provided by the Hololens.

By combining the above two transformations, the transformations from the robot to a hologram can be computed by $_{i}^{r}T = _{h}^{r}T _{i}^{h}T$. The transformations and the coordination of human agents, robots, and virtual holograms are represented in the same coordinate for easy retrieval by the robot.

\subsection{Augmenting Holograms} 

Only knowing the existence of holograms is insufficient; the robot ought to ``see'' the holograms in a way that can be naturally processed for its internal modules (\eg, planning, reconstruction). We design a rendering schema to ``augment'' holograms to the robot and incorporate them into the robot's ROS data messages, such as 3D point clouds and 2D images.

\paragraph*{3D Point Clouds}
The holograms rendered for human agents are stored in a mesh format. To render them in 3D for robots, we use a sampling-based method~\cite{rusu20113d} to convert holograms to point clouds. With the established transformations, these holograms are augmented to the robot's point clouds input with both position and color information; see \cref{fig:exp_env} for examples of rendered holograms for the robot.

\paragraph*{2D Image Projection}
We render the holograms by projecting them onto the robot's received images. Following a general rendering pipeline~\cite{angel2001interactive}, we retrieve the hologram's coordinate $P_c $ with respect to camera frame by the established transformation $_{i}^{r}T$ and calculate the 2D pixel position $P_s = M_p \times P_c$, given the camera's intrinsic matrix $M_p$.

\begin{figure}[t!]
    \centering
    \includegraphics[width=0.95\linewidth]{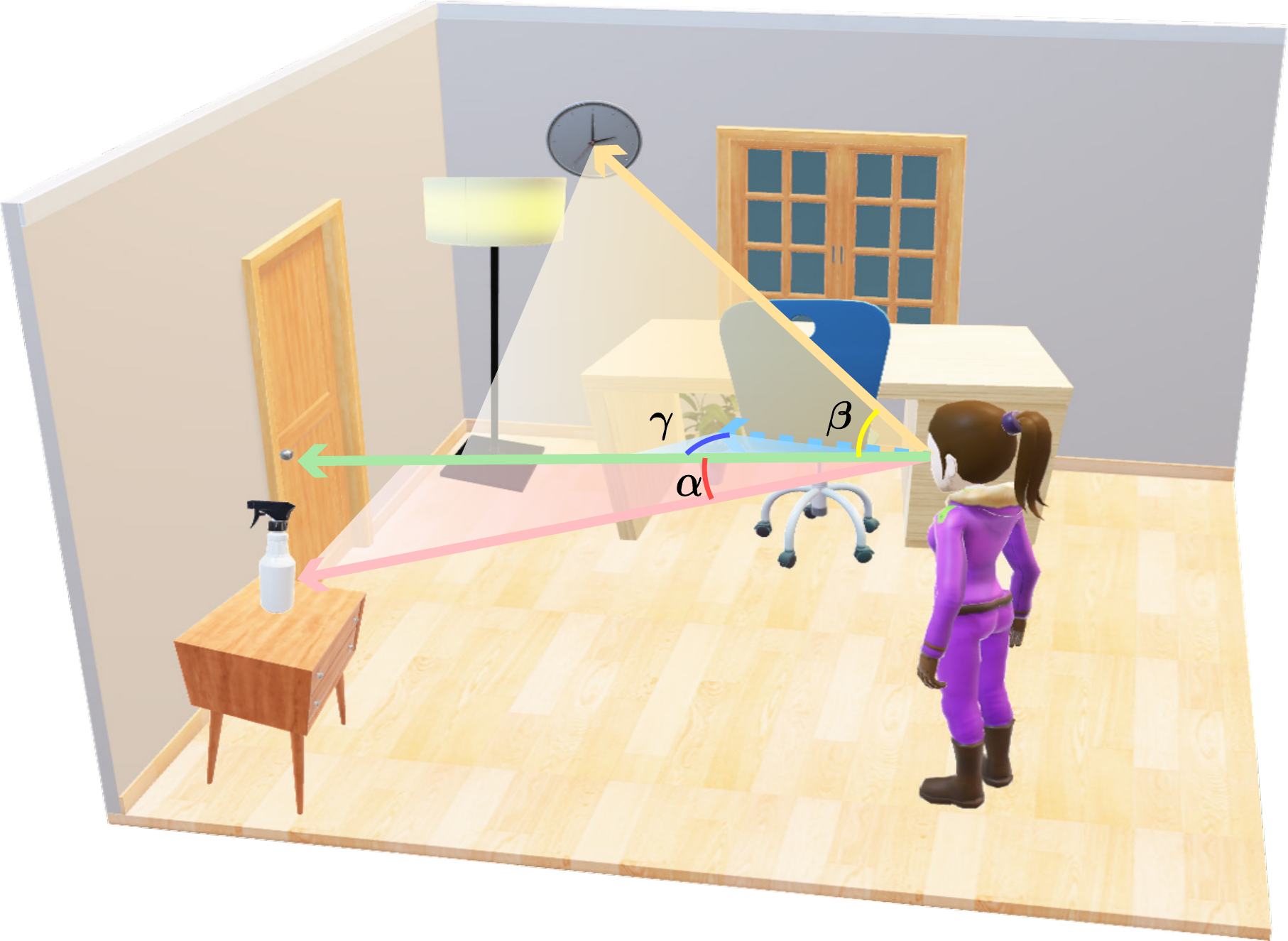}
    \caption{\textbf{The cost of a human agent seeing an object is defined by visibility/occlusion and the angle between two vectors}---her current facing direction and the looking direction of an object. Suppose she is facing to the doorknob (green arrow), the cost to see the clock is higher than seeing the sprayer as the angle $\beta$ is larger than $\alpha$. Although the angle $\gamma$ to see the plant under the desk is smaller than $\beta$, the plant is currently occluded from the human agent's view, resulting in a higher cost despite a smaller angle.}
    \label{fig:visual_cost}
\end{figure}

\subsection{Visual Perspective Taking}\label{sec:vpt} 

Simply ``knowing'' and ``seeing'' holograms would not be sufficient for a robot to help the human agent in the \sarw{} proactively. Instead, to collaborate, plan, and execute properly, the robot would need to possess the ability to infer whether \emph{others} can see an object. Such an ability to attribute others' perspective is known as \acf{vpt}~\cite{hamilton2009visual,lempers1977development}. Specifically, we hope to endow the robot in the \sarw{} with capabilities of inferring (i) whether the human agent can see certain objects, and (ii) how difficult it is.

\ac{vpt} of a robot is devised and implemented at both the object level and scene level.
At the object level, we define the human agent's cost to see an object as a function proportional to the angle between the human agent's current facing direction and looking direction of the object; see an illustration in \cref{fig:visual_cost}. The facing direction is jointly determined by the pose detection from the robot's view and the IMU embedded in HoloLens.
The system also accounts for the visibility of objects as they may be occluded by other real/virtual objects in the environment. To identify an occluded object, multiple virtual rays are emitted from \ac{ar} headset's \ac{fov} to the points in a standard plane whose pose would be updated along with the human agent's pose. The object would be identified as occluded if any of those rays intersect with (i) other holograms whose poses are known in the system, or (ii) real objects or structures whose surfaces are detected by HoloLens's spatial mapping.

At the scene level, we categorize the augmented environment into three regions: (i) \emph{Focusing region}, highlighted in yellow in \cref{fig:system_diagram}e, is considered within the human's \ac{fov} excluding occluded regions, determined by the \ac{fov} of HoloLens---a $30^\circ$ by $17.5^\circ$ area centered at the human's eye. (ii) \emph{Transition region}, highlighted in light blue, does not directly appear in the human's \ac{fov}, but it can be perceived with minimal efforts (\eg, by turning head). (iii) \emph{Blocked region}, highlighted in the dark blue, is occluded and cannot be seen by merely rotating view angles; the human agent has to traverse the space with large body movements, \eg, spaces under tables are typical \emph{Blocked regions}.

\subsection{Interacting with Holograms}

By ``seeing,'' ``knowing,'' and even ``inferring'' human agents about holograms in the \sarw{}, the robot could subsequently plan and manipulate these holograms as an active user in the very same way as a human agent does. However, the holograms are not yet tangible for the robot to ``interact.'' In our prototype system, we devise a simple rule-based algorithm to determine the conditions to be triggered for a robot to interact with holograms.

\begin{figure}[t!]
    \centering
    \includegraphics[width=0.8\linewidth]{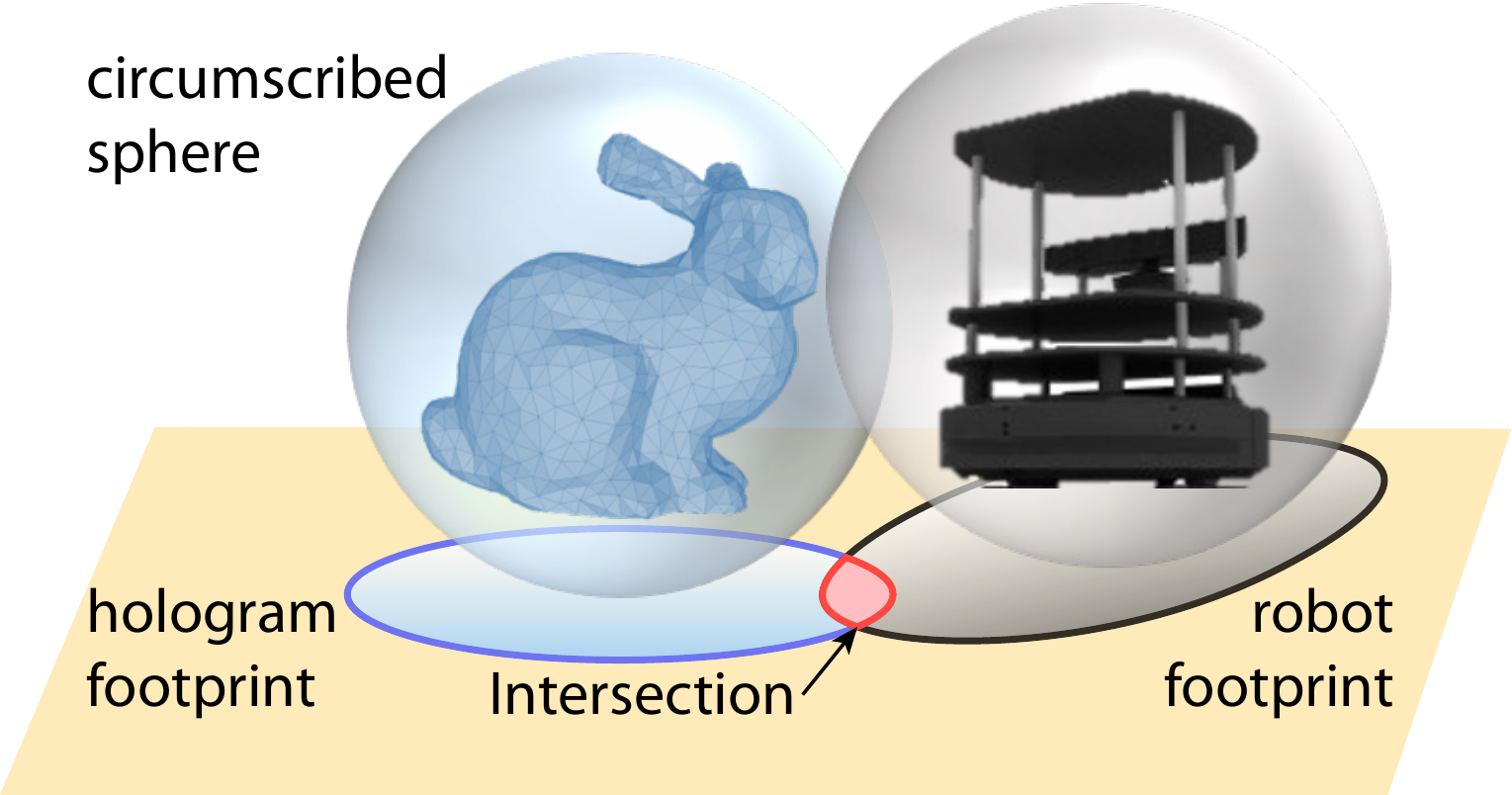}
    \caption{\textbf{Interactions with holograms.} When approaching the hologram, the robot can manipulate a hologram and carry it together if there exists an intersection between their footprints, which triggers the interaction into a manipulation mode.}
    \label{fig:footprint}
\end{figure}

\cref{fig:footprint} illustrates the core idea. After obtaining a hologram's 3D mesh, the algorithm fits a circumscribed sphere to the mesh and to itself with $20\%$ enlargement. Once the robot's sphere is sufficiently close to the hologram's (\ie, there is an intersection between two spheres), it triggers a manipulation mode, and the hologram is attached to the robot and move together. The movements are also synced in the shared perception to the human agent in real-time; see \cref{fig:system_diagram}f. Since the present study adopts a ground mobile robot, we project the spheres to circles on the floor plane to simplify the intersection check. More sophisticated interactions, such as a mobile manipulator grasping a hologram in 3D space, is achievable using standard collision checking methods.

\subsection{Planning} 

The last component of the system is the planner. In fact, the \sarw{} poses no constraints on task and motion planning algorithms; the decision should be made mainly based on robot platforms (\eg, ground mobile robot, mobile manipulator, humanoid) and executed tasks (\eg, \ac{hri}, navigation, prediction) during the interactions; see the next section for the planning schema adopted in this paper. 

\section{Experiment}\label{sec:exp}

\subsection{Experimental Setup}

\begin{figure}[t!]
    \centering
    \begin{subfigure}[b]{\linewidth}
        \includegraphics[width=\linewidth]{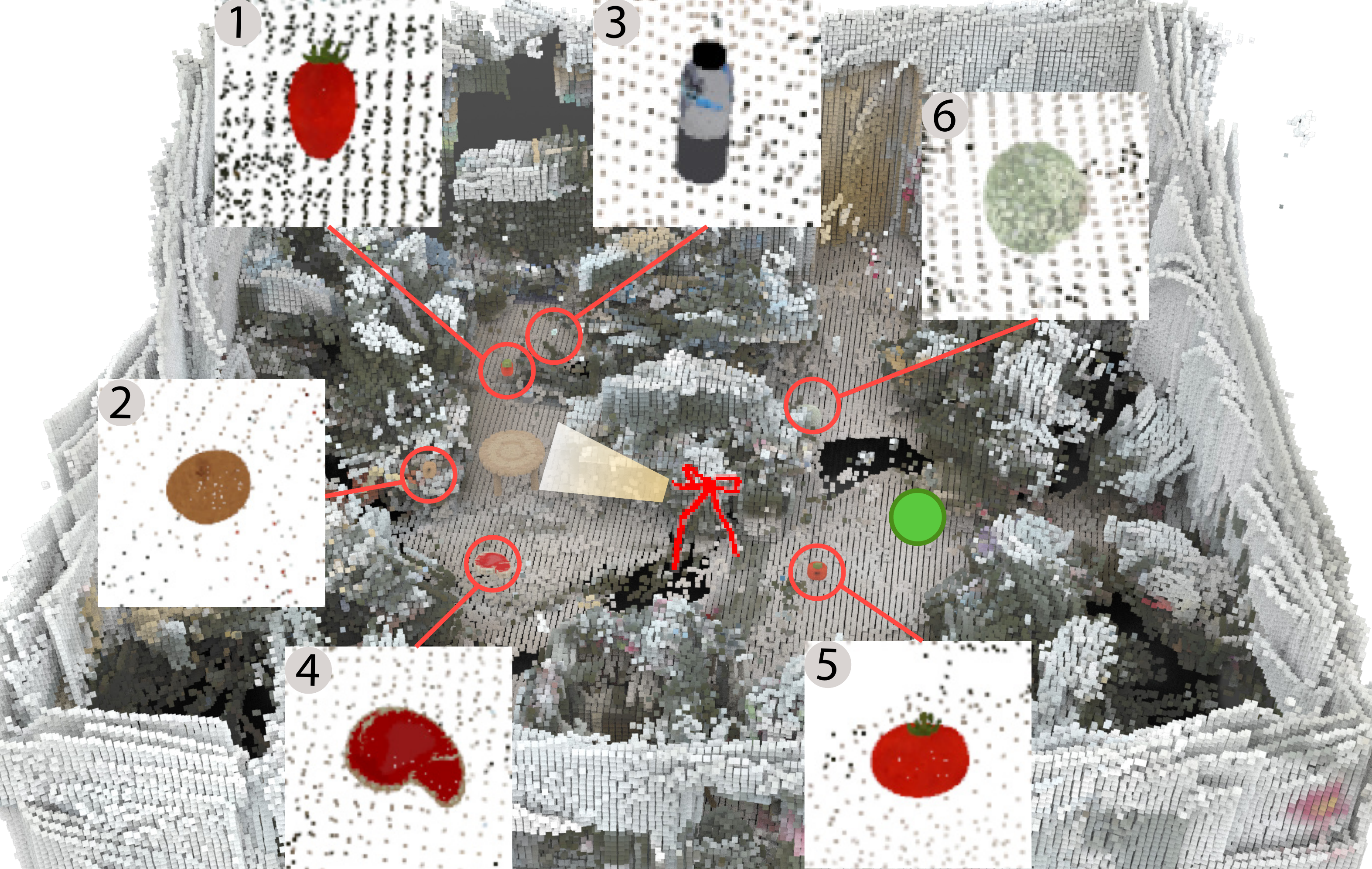}
        \caption{Experimental environment}
        \label{fig:exp_env}
    \end{subfigure}%
    \\
    \begin{subfigure}[b]{\linewidth}
        \includegraphics[width=\linewidth]{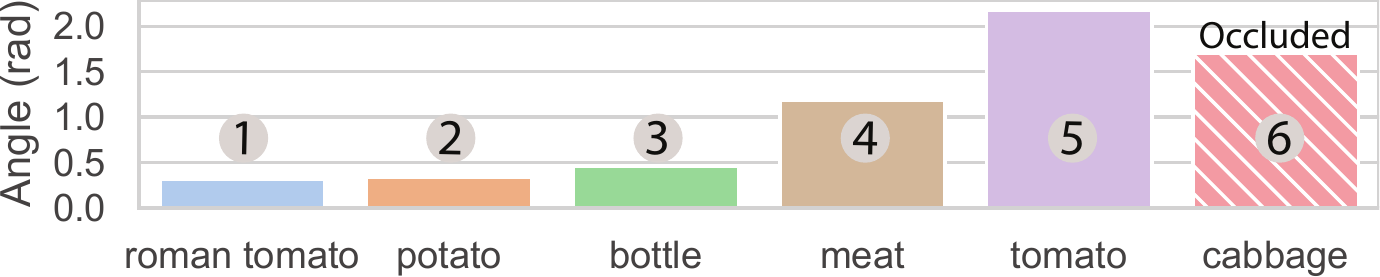}
        \caption{Costs of seeing various holograms}
        \label{fig:object_cost}
    \end{subfigure}%
    \caption{\textbf{Environment and estimated costs.} (a) The experimental environment rendered as point clouds from the robot's view. The red skeleton is the detected human pose, the yellow area the human facing direction, and the green dot the robot's initial position. (b) Human agent's cost of seeing holograms with object 6 occluded.}
\end{figure}

We design a resource collection game in the \sarw{} to demonstrate the efficacy of the system. \cref{fig:exp_env} depicts the environment. Six holograms, rendered as point clouds and highlighted in circles with zoomed-in views, are placed around the human agent (marked by a red skeleton at the center of the room), whose facing direction is indicated in yellow. Some holograms can be easily seen, whereas others are harder due to their tricky locations in 3D or occlusion (\eg, object 6). A human agent's task is to collect all holograms and move them to the table as fast as possible. The robot stationed in the green dot would help the human in collecting the resources. 

As described in \cref{sec:vpt}, the robot first estimates the cost for a human agent to see the holograms and whether they are occluded; the result is shown in \cref{fig:object_cost}. In our prototype system, the robot prioritizes to help the occluded holograms and then switch to the one with the highest cost. In future, it is possible to integrate prediction models (\eg, \cite{hoffman2007cost,grigore2018preference,qi2020generalized}) that anticipate human behaviors.

\subsection{Qualitative Results}

Intuitively, we should see a better overall performance during \ac{hri} via \sarw{} due to its shared perception and manipulation that enables a robot to help the human agent for task completion collaboratively proactively.

\begin{figure*}[t!]
    \centering
    \begin{subfigure}[b]{\linewidth}
        \includegraphics[width=\linewidth]{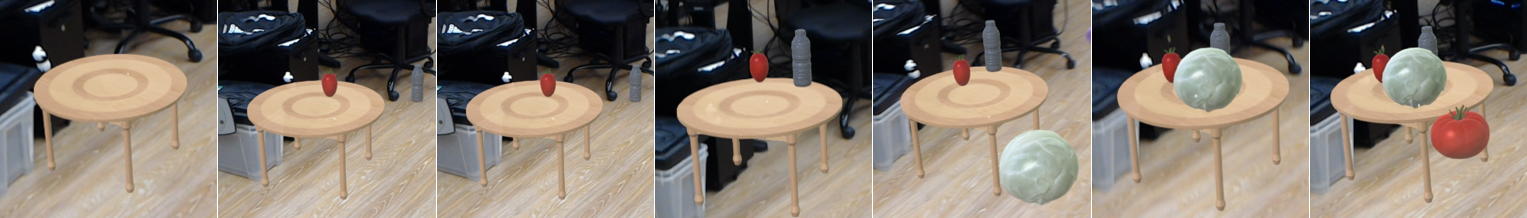}
        \caption{Human agent's egocentric view captured by the HoloLens.}
    \end{subfigure}%
    \\
    \begin{subfigure}[b]{\linewidth}
        \includegraphics[width=\linewidth]{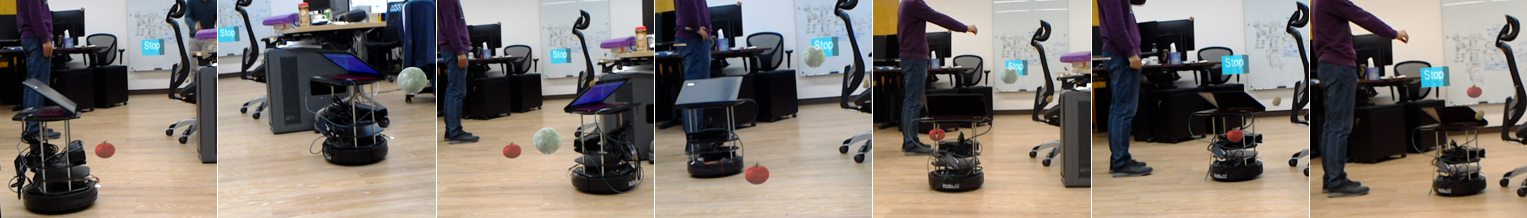}
        \caption{Third-person view of robot motions.}
    \end{subfigure}%
    \\
    \begin{subfigure}[b]{\linewidth}
        \includegraphics[width=\linewidth]{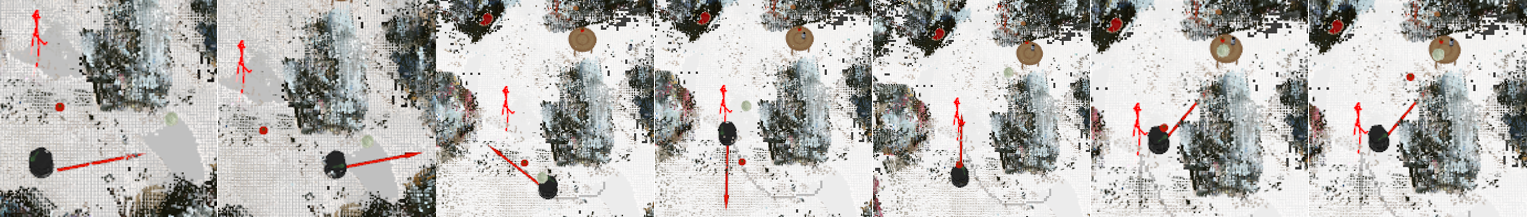}
        \caption{Robot's knowledge and plans.}
    \end{subfigure}%
    \caption{\textbf{Qualitative results.} Qualitative experimental results in the resource collecting game. The robot helps to collect holograms (object 5 and 6 in \cref{fig:object_cost}) that are difficult for the human agent to see.}
    \label{fig:qualitative}
\end{figure*}

\cref{fig:qualitative} gives an example of a complete process, demonstrating a natural interaction between the human agent and the robot to accomplish a given task collaboratively. The top row shows the human agent's egocentric views through the Hololens that overlays the holograms to the image captured by its PV camera. The middle row is a sequence of the interactions between the robot and holograms from a third-person view. The bottom row reveals the robot's knowledge of the workspace and its plans. In this particular trial, the human agent first collected the roman tomato and the bottle as they appear to have a lower cost. In parallel, the robot collaboratively carries holograms---the occluded cabbage and the tomato with the highest cost---to the human agent.

\subsection{Quantitative Results}

We conduct a pilot study to evaluate \sarw{} quantitatively. Twenty participants were recruited to assess the robot performance in a between-subject setting ($N=10$ for each group). The participants in the \emph{Human} group are asked to find and collect all six holograms by themselves. The participants in the \emph{Human+Robot} group use the \sarw{} system, where the robot proactively helps the participants to accomplish the task. Each subject has no familiarization with the physical environments, but they received simple training about how to use the \ac{ar} device right before the experiments started.

\begin{figure}[t!]
    \centering
    \includegraphics[width=\linewidth]{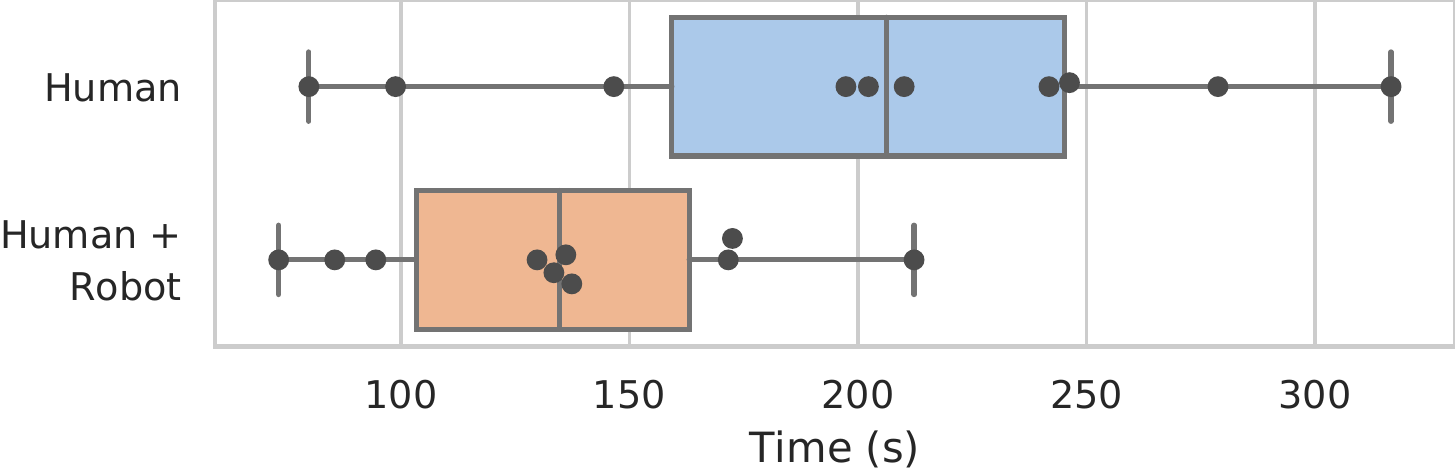}
    \caption{\textbf{Quantitative results.} Box plot of all participants' collection time in two different groups; the dots in the plot are the individual data points. Subjects helped by the robot in the \sarw{} are significantly more efficient in task completion.}
    \label{fig:quanlitative}
\end{figure}

\cref{fig:quanlitative} compares the results between the two groups. The difference of the completion time is statistically significant; $t(19)=1.0$, $p=0.028$. Participants with robot's help take significantly less time (mean: $135$ seconds, median: $134$ seconds) to complete the given task. In contrast, the baseline group requires much more time with a larger variance (mean: $202$ seconds, median: $206$ seconds). This finding indicates a new role that a robot can play in the \sarw{} by assisting human agents to accomplish a task collaboratively.

\setstretch{0.93}

\section{Related Work and Discussion}\label{sec:related_work}

We design, implement, and demonstrate how the shared perception and manipulation provided by the \sarw{} improve \ac{hri} with a proof-of-concept system using a resource collection game. In future, more complex and diverse \ac{hri} studies are needed to further examine and benifits and limits of the \sarw{} by (i) varying the degree of human agent's and/or robot's perception and manipulation capability; \eg, only the robot can see and act on holograms while the human agent cannot, as an opposite to current \ac{ar} setup, and (ii) introducing virtual components to avoid certain costly and dangerous setups in the physical world. Below, we briefly review related work and scenarios that \sarw{} could potentially facilitate.

The idea of creating a \textbf{shared workspace} for human agents and robots has been implemented in \ac{vr}, where they can re-target views to each other to interact with virtual objects~\cite{churchill1998collaborative}. Prior studies have demonstrated advantages in teleoperation~\cite{lipton2017baxter} and robot policy learning~\cite{zhang2018deep}. More recently, a system~\cite{grandi2017design,zhang2018cars} that allows multiple users to interact with the same \ac{ar} elements is devised. In comparison, the \sarw{} deals with the perceptual noise in the physical world and promotes robots to become active users in \ac{ar} to work on tasks with humans collaboratively.

In recent years, \textbf{Human-Robot Interaction and Collaboration} have been developing with increasing breadth and depth. One core challenge of the field is to seek how the robot or the human should act to promote understanding and trust, usually in terms of predictability, with the other. From a robot's angle, it models humans by inferring goals~\cite{liu2016goal,pellegrinelli2016human}, tracking mental states~\cite{devin2016implemented,yuan2020joint}, predicting actions~\cite{unhelkar2018human}, and recognizing intention and attention~\cite{pandey2010mightability,huang2016anticipatory}. From a human agent's perspective, the robot needs to be more expressed~\cite{zhou2018cost}, to promote human trust~\cite{edmonds2019tale}, to assist properly~\cite{kato2015may,mollaret2016multi}, and to generate proper explanations of its behavior~\cite{edmonds2019tale}. We believe the proposed \sarw{} is an ideal platform for evaluating and benchmarking existing and new algorithms and models.

\textbf{Human-robot teaming}~\cite{gombolay2017computational,talamadupula2010planning} poses new challenges to computational models aiming to endow robots with the Theory of Mind abilities, which are usually in a dyadic scenario~\cite{premack1978does}. With the adaptability to multi-party settings and the fine-grained controllability of users' situational awareness, the proposed \sarw{} offers a unique solution to test the robot's ability to maintain belief, intention, and desires~\cite{holtzen2016represent,wei2018and,yuan2020joint} of other agents. Crucially, the robot would play the role of a collaborator to help and as a moderator~\cite{short2017robot} to accommodate each agent. The ultimate goal is to forge a shared agency~\cite{kleiman2016coordinate,ho2016feature,tang2020bootstrapping,stacy2020intuitive} between robots and human agents for seamless collaboration.

How human's \textbf{cognition} emerges and develops is a fundamental question. Researchers have looked into the behaviors of primates' collaboration and communication~\cite{melis2019chimpanzees}, imitation~\cite{gallese1996action}, and crows' high-level reasoning~\cite{taylor2009new}, planning and tool making~\cite{hunt1996manufacture} for deeper insights. Cognitive robots are still in their infancy in developing such advanced cognitive capabilities, despite various research efforts~\cite{deng2019embodiment,zhu2020dark}. These experimental settings can be relatively easier to replicate in the \sarw{}, which would open up new avenues to study how a robot would emerge similar behaviors.

\setstretch{1}
\balance
\bibliographystyle{ieeetr}
\bibliography{IEEEfull}
\end{document}